\documentclass{article}
\usepackage{spconf,amsmath,graphicx}
\usepackage{setspace}

\usepackage{graphicx} %add figure
\usepackage{booktabs}
\usepackage{cite}
\usepackage{subfigure} % Í¼Ïñ
\usepackage{color}
\usepackage{epstopdf}
\usepackage{multirow}
\usepackage{color,xcolor}
\usepackage{amsmath}
\usepackage{url}
\usepackage{algorithmic}
\usepackage{algorithm}

\small
\title{Towards General Deepfake Detection with Dynamic Curriculum}

\name{Wentang Song\thanks{This work was supported by NSFC (Grant U23B2022 and U22B2047), Guangdong Basic and Applied Basic Research Foundation (Grant 2019B151502001), Shenzhen R\&D Program (Grant JCYJ20200109105008228). Bin Li is the corresponding author}, Yuzhen Lin, Bin Li$^{\star}$}
\address{Guangdong Key Laboratory of Intelligent Information Processing, Shenzhen Key Laboratory
	\\of Media Security, Shenzhen University, Shenzhen 518060, China}

\begin{document}
	%\ninept
	%

\maketitle

\begin{abstract}
Most previous deepfake detection methods bent their efforts to discriminate artifacts by end-to-end training. However, the learned networks often fail to mine the general face forgery information efficiently due to ignoring the data hardness. In this work, we propose to introduce the sample hardness into the training of deepfake detectors via the curriculum learning paradigm. Specifically, we present a novel simple yet effective  strategy, named Dynamic Facial Forensic Curriculum (DFFC), which makes the model gradually focus on hard samples during the training. Firstly, we propose Dynamic Forensic Hardness (DFH) which integrates the facial quality score and instantaneous instance loss to dynamically measure sample hardness during the training. Furthermore, we present a pacing function to control the data subsets from easy to hard throughout the training process based on DFH. Comprehensive experiments show that DFFC can improve both within- and cross-dataset performance of various kinds of end-to-end deepfake detectors through a plug-and-play approach. It indicates that DFFC can help deepfake detectors learn general forgery discriminative features by effectively exploiting the information from hard samples.
\end{abstract}

\section{Introduction}
Deepfake techniques \cite{FSGANSubjectAgnostic2019nirkin,AdvancingHighFidelity2020li,FSGANv2ImprovedSubject2023nirkin} refer to a series of deep learning-based facial forgery techniques that can swap or reenact the face of one person in a video to another. They can lead to the dissemination of false information or even political manipulation.
Thus, detecting deepfakes has become a crucial research topic in recent years. 

Early works\cite{MesoNetCompactFacial2018afchar,FaceForensicsLearningDetect2019rossler,ThinkingFrequencyFace2020qian,MultiAttentionalDeepfakeDetection2021zhao} treat deepfake detection as a binary classification problem, and commonly use deep neural networks to distinguish fake faces. In order to improve the detection performance, some works \cite{SpatialPhaseShallowLearning2021liu,GeneralizingFaceForgery2021luo,ExposingFaceForgery2022chen} introduce auxiliary modalities or supervision  information for learning subtle forgery artifacts.  However, these methods can not mine the general face forgery information efficiently due to ignoring the data hardness, resulting in poor generalization performance under the real-world scenario.
To address this issue, some works introduce pseudo-forgery augmentations\cite{FaceXRayMore2020li,LearningSelfConsistencyDeepfake2021zhao,DetectingDeepfakesSelfBlended2022shiohara,LearningFeaturesIntraConsistency2023chen} to enlarge the diversity of forgery artifacts, but are prone to fail in detecting samples with heavy post-processing.
%However, these methods require the expert knowledge and non-universal modules to enhance the performance. However, these methods can not thus achieve promising generalization performance, which is one of the major concerns for existing deepfake detection systems. Thus, most recent works are dedicated to improving the generalization ability of deepfake detectors.
\begin{figure}[t]
	\centering
	\subfigure[real faces]{
		\includegraphics[width=0.47\linewidth]{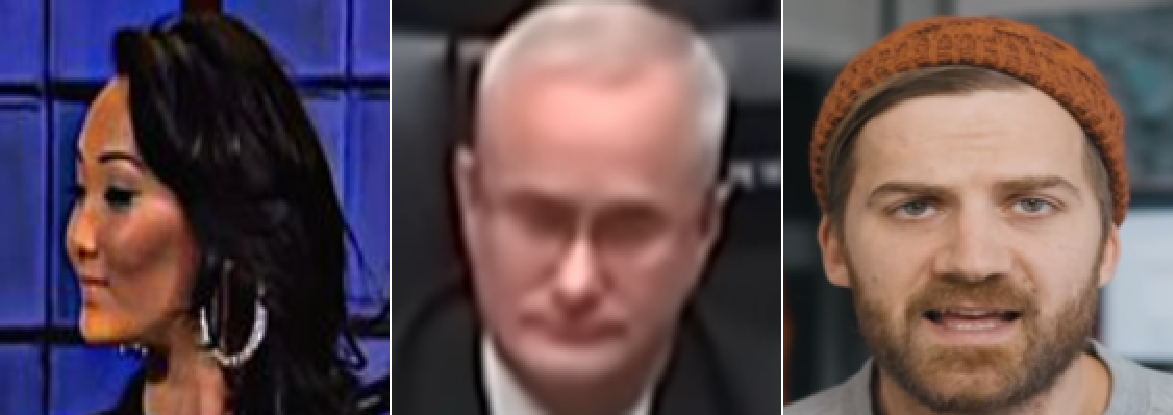}
		\label{fig:example1}
	}
	\subfigure[fake faces]{
		\includegraphics[width=0.47\linewidth]{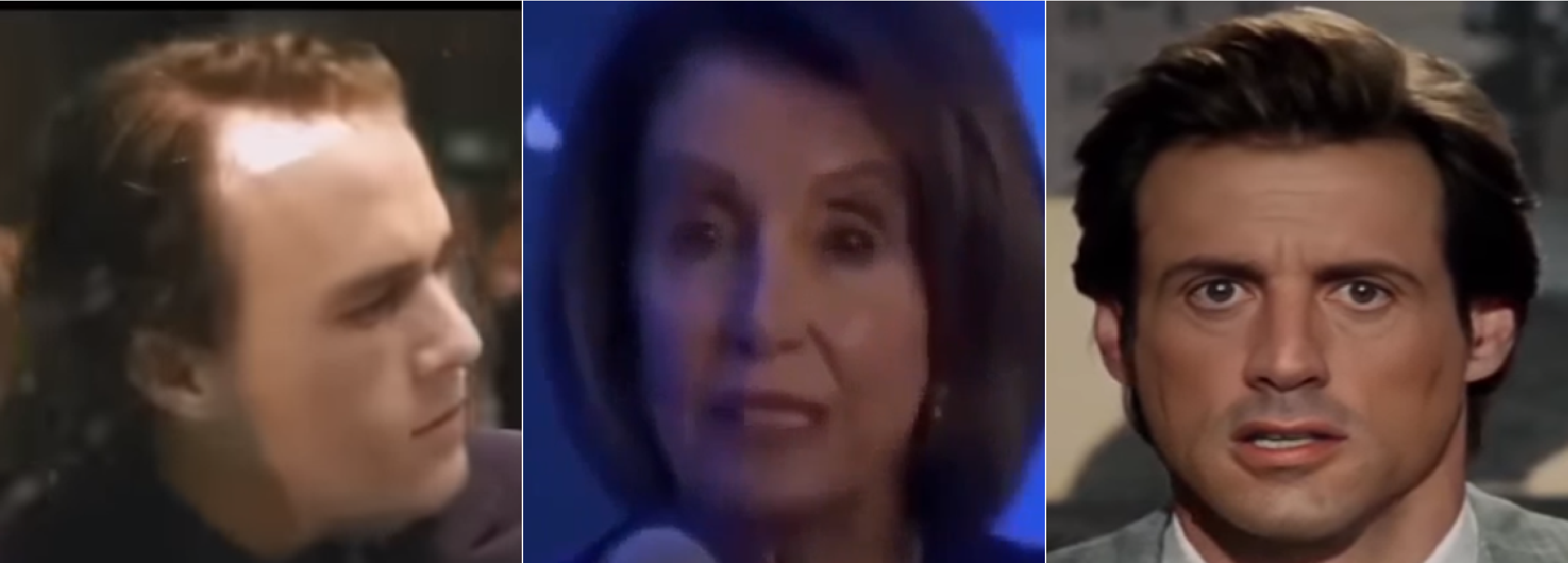}
		\label{fig:example2}
	}
	\caption{Real and fake faces with various visual qualities on YouTube.}
	\label{fig:examples}
\end{figure}

As shown in Figure \ref{fig:examples}, the judgment of real/fake facial images/videos by human vision is difficult due to the visual qualities.
Take the post-processing for example, the post-processing on real faces can be misinterpreted as forgery artifacts. Besides, the post-processing on fake faces can erase the forgery artifacts. 
By analogy, the discriminative deepfake detectors vary with different samples.
However, most existing detection methods treat all samples the same during the training. Thus, we argue that the hardness of samples should be taken into account for training the general deepfake detector.

Curriculum learning (CL)\cite{CurriculumLearning2009bengio,SurveyCurriculumLearning2022wang,CurriculumLearningSurvey2022soviany}, which involves presenting training samples to the model in a specific order of hardness, is an effective scheme for hard sample mining \cite{AdaptiveCurriculumLearning2021kong,EfficientTrainExploringGeneralized2023wang,AudioSteganalysisImproved2019lin}. 
Inspired by such a paradigm, we present a novel simple yet effective strategy, named Dynamic Facial Forensic Curriculum (DFFC) that makes the model gradually focus on hard samples during the training. We introduce Dynamic Forensic Hardness (DFH), a novel approach that integrates the facial quality score and instantaneous instance loss to dynamically assess sample hardness throughout the training phase.
Additionally, we present a pacing function that guides the progression of training subsets from easy to hard based on the DFH score throughout the training iterations.
Experimental results demonstrate that DFFC can improve both within- and cross-dataset performance of various kinds of deepfake detectors through a plug-and-play approach.
The main contributions of our work are summarized as follows:

\begin{itemize}
	\item To the best of our knowledge, the proposed DFFC is the first work that introduces the curriculum learning paradigm to mine hard samples for the deepfake detection task. DFFC can be plug-and-plays for any end-to-end deepfake detectors.
	\item To measure the sample hardness, we propose the DFH score that integrates the instantaneous instance loss and facial quality score dynamically during the training. 
	\item To adequately mine the forgery information on hard samples, we present a pacing function to gradually control the training subsets from easy to hard according to DFH throughout the training process. 
%	To the best of our knowledge, this is the first work to introduce hard sample mining into deepfake detection.
%	\item Experimental results demonstrate that DFFC can improve both within- and cross-datasets performances of various kinds of deepfake detectors.
\end{itemize}

\section{Methodology}
\subsection{Overview}
\begin{figure}[t]
	\centering
	\includegraphics[width=\linewidth]{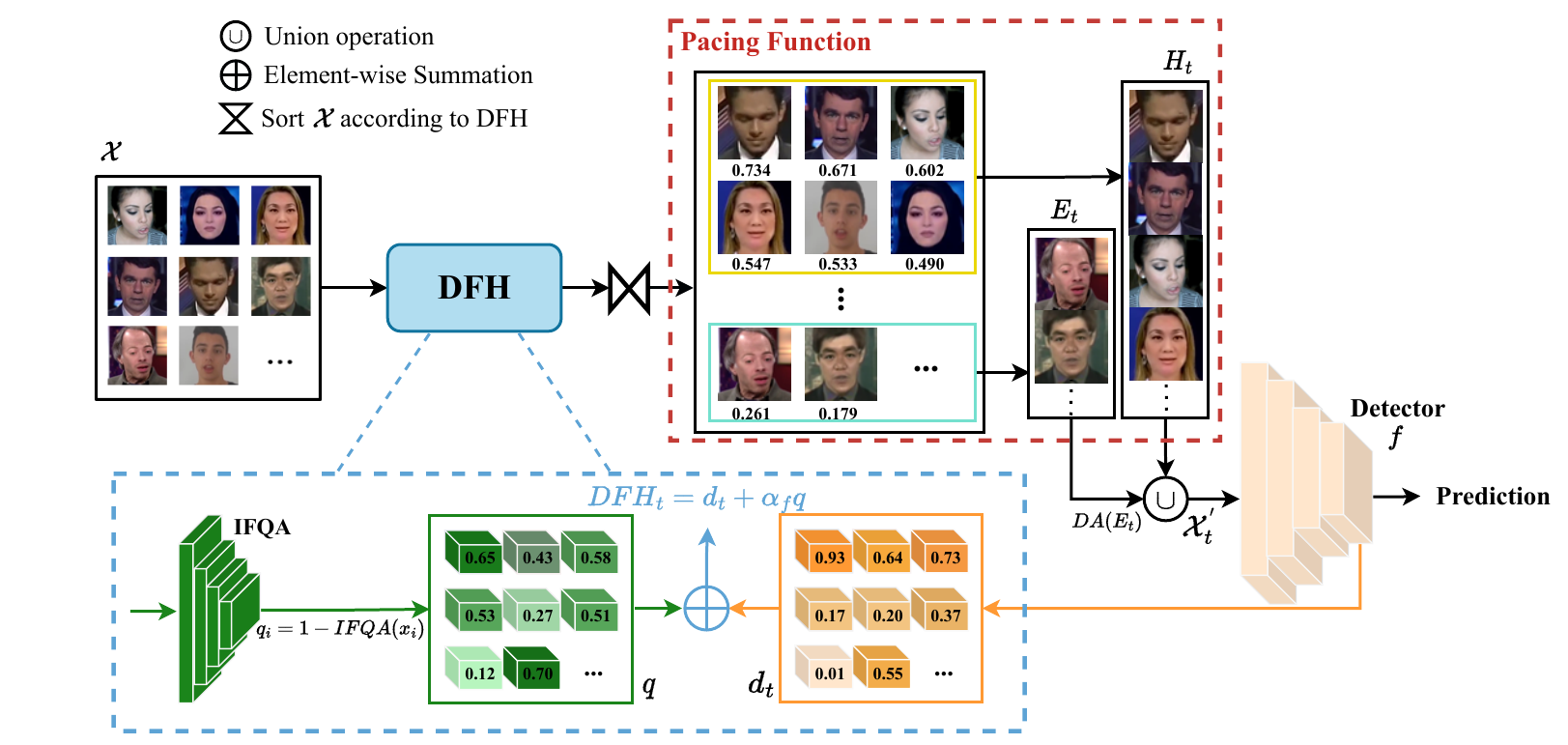}
	\caption{The overall pipeline of the proposed DFFC. }
	\label{fig:overview}
\end{figure}

In this section, we introduce the proposed Dynamic Facial Forensics Curriculum (DFFC). Specifically, the overall pipeline (See in Figure \ref{fig:overview}) mainly consists of two components: \textit{1) Dynamic Forensic Hardness (DFH)} is proposed to measure the sample hardness during the training; \textit{2) Pacing Function} is presented to control the pace of presenting data from easy to hard according to DFH. Each component will be detailed subsequently.
%The details of each component will be detailed subsequently. which are arranged in the order of increasing average difficulty score. We use the pre-trained model to obtain the initial difficulty score s0 and then gradually adjust the difficulty score using the current losses of examples. With the difficulty score, we sort the full dataset X in ascending order, and obtain a sample pool X ′ which contains the first p(·) easier examples. Then the next minibatch will be randomly selected from the sample pool X ′. In our method, we use pacing functions p(·) that increase with iterations, so the average difficulty score of mini-batch also tends to increase with iterations, meaning that increasing numbers of “hard” examples are added to the sample pool X ′. The pseudo-code of the proposed algorithm is given in Algorithm

\subsection{Dynamic Forensic Hardness}
The difficulty score plays a key role in curriculum learning as it describes the relative “hardness” of each sample. In this work, we propose Dynamic Forensic Hardness (DFH), which considers the dynamic model behavior and the static facial quality. 
Let $ \mathcal{X} = {(x_{i},y_{i} )}_{i=1}^{N} $ be the training dataset with $N$ samples, $x_{i}$ and $y_{i}$ represent the $i$-th data and its ground-truth label, respectively. Let $f(\cdot,\theta_{t})$ be the deepfake detector model with the parameter $\theta_{t}$ at $t$-th epoch. 
We regard the loss (i.e., the binary entropy loss denoted as $l(f(x_{i}, \theta_{t-1}), y_{i})$) of $i$-th sample at $t-1$-th epoch as an indicator for the current hardness of this sample judged by the current state of the model before conducting the training step.  Thus, we propose the instantaneous hardness $s_{t}(i)$ that normalized the current loss with the learning rate $\eta_{t}$, formulated as:
\begin{equation}\label{eq1}
	s_{t}(i)=l(f(x_{i};\theta_{t-1} ),y_{i})\cdot \eta_{max}/\eta_{t},
\end{equation}
where $\eta_{max}$ is the max learning rate during the training. 
Inspired by Dynamic Instance Hardness (DIH)\cite{CurriculumLearningDynamic2020zhou}, we measure the dynamic hardness ${d}_{t}(i)$ through a moving average of the instantaneous hardness ${s}_{t}(i)$ over training history, defined and computed recursively as:
\begin{equation}\label{eq2}
	{d}_{t}(i)=
	\begin{cases}
		\gamma \times s_{t}(i) + (1-\gamma) \times {d}_{t-1}(i),& \text{if} \ \, i\in H_t\\
		{d}_{t-1}(i),& \text{otherwise},
	\end{cases}
\end{equation}
where $\gamma \in [0,1]$ is a discount factor, and $H_t$ is the subsets of hard samples at $t$-th epoch selected by the pacing function.

As mentioned in Section 1, detecting low-quality faces is harder than high-quality ones for deepfake detectors. We regard that the facial visual quality can be a prior hardness for deepfake detection tasks. To achieve this, we utilize a pre-trained facial quality assessment model IFQA \cite{IFQAInterpretableFace2023jo} as a teacher to guide the prior hardness as $ q(i)=1-IFQA(x_i)$. Finally, we get DFH by integrating $d_{t}(i)$ and $q(i)$ as: 
\begin{equation}\label{DFH}
    DFH_{t}(i)={d}_{t}(i)+\alpha_{f}q(i)
\end{equation}
where $\alpha_{f}$ is a balance weight. 

%We also introduce the facial quality score $q(i)$. The facial quality score of difficult samples and simple samples screened by DIH is significantly different, which means to be a good guidance at the beginning of training. 

\begin{algorithm}[t]  %其中这里面不能有H不然会报错，不过不影响结果
	\caption{Pacing Function}%算法名字
	\begin{algorithmic}[1]
		\REQUIRE $ \mathcal{X},t, DFH_{t}(i),\alpha_{f}, \mathbf{T}=[T_{0}, \dots , T_{N}], \mathcal{DA}(\cdot), \alpha_{k}$
		\ENSURE Sample pool $\mathcal{X}^{\prime}_{t}$ in $t$-th epoch%输出
		
%		Initialize $k$
		\IF{$t\le T_{0}$}
		\STATE $\mathcal{X}^{\prime}_{t}\gets \mathcal{X}, k_{0}=N$
		\ELSE 
		\IF{$t\in \mathbf{T}$} 
		\STATE $k_{n}\gets k_{n-1} \times \alpha_{k}$
%		 $\alpha_{f}\gets \beta_{f}\times \alpha_{f}$
		\ENDIF
		
%		 Compute $DFH_{t}(i)$ with Eq (\ref{DFH})
		
		$H_{t}\gets argmax_{\mid H_{t} \mid= k_{n}}\sum_{i\in \mathcal{X}}DFH_{t}(i)$
		
		$E_{t}\gets argmin_{\mid E_{t} \mid= E}\sum_{i\in \mathcal{X}}DFH_{t}(i)$
		
		$\mathcal{X}^{\prime}_{t}\gets H_{t}\cup \mathcal{DA}(E_{t})$
		\ENDIF

	\end{algorithmic}
\end{algorithm}

\subsection{Pacing Function}
To control the learning pace of presenting data from easy to hard, we design a pacing function to determine the sample pool $\mathcal{X}^{\prime}_{t}$ of training data according to DFH. The pipeline is summarized in Algorithm 1. 
Like human education, if a teacher presents materials from easy to hard in a very short period of time, students would become confused and will not learn effectively.
Thus, we define a pacing sequence $\mathbf{T}=[T_{0}, \dots , T_{N}]$ to represent $n+1$ milestones(i.e., $n$ episodes) in total training epoch $T$.
During the first $T_{0}$ epochs, we utilize all the samples in $\mathcal{X}$ for the warm-up training. After epoch $T_{0}$, we only change the size of $\mathcal{X}^{\prime}_{t}$ at every milestone $T_{n}$. Specifically, at each epoch $t$ of episode $n$, we select $k_{n}$ samples with top DFH in $\mathcal{X}$ (i.e., hardest samples) as a hard sample pool $H_{t}$. Along with the training, we reduce the size of $H_{t}$ by $k_{n}\gets \alpha _{k}\times k_{n-1}$ with discount factor $\alpha_{k}$ to make it gradually focus on harder samples. To further enlarge the diversity of the data, we also select the $E$ samples with bottom DFH in $\mathcal{X}$ (i.e., easiest samples) as a easy sample pool $E_{t}$ and then conduct lightweight data augmentations $DA(\cdot)$ (e.g., Gaussian blur, brightness adjustment, and affine transformation) on them.
Finally, we get the sample pool $\mathcal{X}^{\prime}_{t}$ by mixing the $H_{t}$ and $DA(E_{t})$, i.e, $\mathcal{X}^{\prime}_{t}\gets H_{t}\cup DA(E_{t})$. After that, we sample mini-batch in $\mathcal{X}^{\prime}_t$ and then conduct a vanilla training step to update the model parameter as $\theta_{t}$.

%supervised learning paradigm.
\begin{table}[t]
	\centering
	\caption{Frame-level ACC(\%) for detecting five manipulations on FF++ (LQ). The best results are highlighted.}
	\setstretch{0.85}
	\resizebox{\linewidth}{!}{
		\begin{tabular}{ccccccc}
			\toprule
			\multicolumn{2}{c}{{Methods}} & DF    & F2F   & FS    & NT    & FSh \\
			%		\cmidrule{3-7}    \multicolumn{2}{c}{} & ACC   & ACC   & ACC   & ACC   & ACC \\
			\midrule
			\multicolumn{2}{c}{Xception\cite{XceptionDeepLearning2017chollet}} & 95.15  & 83.48  & 92.09  & 77.89  & 94.96  \\
			\multicolumn{2}{c}{+ DIH} & 96.34  & 90.14  & 94.05  & 78.93  & 95.32  \\
			\multicolumn{2}{c}{+ DFFC} & \textbf{96.43} & \textbf{90.88} & \textbf{94.64} & \textbf{79.50} & \textbf{95.36} \\
			\midrule
			\multicolumn{2}{c}{ENb4\cite{EfficientNetRethinkingModel2019tan}} & 94.26  & 86.86  & 92.72  & 75.99  & 94.54  \\
			\multicolumn{2}{c}{+ DIH} & 97.11  & 90.55  & 95.34  & 80.29  & 95.41  \\
			\multicolumn{2}{c}{+ DFFC} & \textbf{97.28} & \textbf{91.22} & \textbf{95.38} & \textbf{80.65} & \textbf{95.59} \\
			\midrule
			\multicolumn{2}{c}{MAT\cite{MultiAttentionalDeepfakeDetection2021zhao}} & 95.36  & 88.36  & 93.21  & 77.50  & 94.62  \\
			\multicolumn{2}{c}{+ DIH} & 97.25  & 90.20  & 95.23  & 78.95  & \textbf{95.45} \\
			\multicolumn{2}{c}{+ DFFC} & \textbf{97.59} & \textbf{90.55} & \textbf{95.66} & \textbf{79.05} & 95.12  \\
			\midrule
			\multicolumn{2}{c}{Swin\cite{SwinTransformerV22022liu}} & 95.10  & 88.57  & 92.00  & 76.29  & 94.17  \\
			\multicolumn{2}{c}{+ DIH} & 96.78  & 90.55  & 95.40  & 80.04  & 95.43  \\
			\multicolumn{2}{c}{+ DFFC} & \textbf{97.37} & \textbf{91.17} & \textbf{95.59} & \textbf{80.33} & \textbf{95.47} \\
			\midrule
			\multicolumn{2}{c}{SPSL\cite{SpatialPhaseShallowLearning2021liu}} & 93.82  & 86.82  & 91.73  & 75.97  & 91.26  \\
			\multicolumn{2}{c}{+ DIH} & 96.10  & 89.42  & 94.32  & 77.59  & 94.37  \\
			\multicolumn{2}{c}{+ DFFC} & \textbf{96.19} & \textbf{89.55} & \textbf{94.57} & \textbf{77.94} & \textbf{94.50} \\
			\midrule
			\multicolumn{2}{c}{SRM\cite{GeneralizingFaceForgery2021luo}} & 94.63  & 87.71  & 91.27  & 76.39  & 93.36  \\
			\multicolumn{2}{c}{+ DIH} & 95.38  & 89.40  & 93.23  & 76.46  & 94.30  \\
			\multicolumn{2}{c}{+ DFFC} & \textbf{96.48} & \textbf{89.92} & \textbf{93.97} & \textbf{77.82} & \textbf{94.89} \\
			\bottomrule
	\end{tabular}}%
	\label{tab:LQ}%
\end{table}%

\begin{table}[t]
	\centering
	\caption{Video-level AUC(\%) on cross-dataset evaluations (trained on FF++/DF (HQ)). The best results are highlighted.}
	\setstretch{0.9}
	\resizebox{\linewidth}{!}{
		\begin{tabular}{ccccccc}
			\toprule
			\multicolumn{2}{c}{{Methods}} & DF & CDF & Wild  & DFDC-P & DFD \\
			%		\cmidrule{3-7}    \multicolumn{2}{c}{} & AUC   & AUC   & AUC   & AUC   & AUC \\
			\midrule
			\multicolumn{2}{c}{Xception\cite{XceptionDeepLearning2017chollet}} & 99.38  & 70.74  & 60.06  & 80.93  & 90.60  \\
			\multicolumn{2}{c}{+ DIH} & 99.71  & 80.45 & 59.91  & 81.15  & 88.68  \\
			\multicolumn{2}{c}{+ DFFC} & \textbf{99.71} & \textbf{82.12} & \textbf{65.87} & \textbf{81.61} & \textbf{93.43} \\
			\midrule
			\multicolumn{2}{c}{ENb4\cite{EfficientNetRethinkingModel2019tan}} & 99.56  & 74.50  & 60.04  & 79.37  & 92.34  \\
			\multicolumn{2}{c}{+ DIH} & 99.58  & 81.19 & 60.43  & 81.68  & 92.00  \\
			\multicolumn{2}{c}{+ DFFC} & \textbf{99.59} & \textbf{85.01} & \textbf{68.56} & \textbf{89.33} & \textbf{95.51} \\
			\midrule
			\multicolumn{2}{c}{MAT\cite{MultiAttentionalDeepfakeDetection2021zhao}} & 99.48  & 76.39  & 61.10  & 76.30  & 92.56  \\
			\multicolumn{2}{c}{+ DIH} & 99.49  & 80.46 & 60.23  & 81.18  & 92.37  \\
			\multicolumn{2}{c}{+ DFFC} & \textbf{99.50} & \textbf{84.37} & \textbf{66.71} & \textbf{82.87} & \textbf{94.73} \\
			\midrule
			\multicolumn{2}{c}{Swin\cite{SwinTransformerV22022liu}} & 99.66  & 73.53  & 69.72  & 88.18  & 93.07  \\
			\multicolumn{2}{c}{+ DIH} & 99.68  & 91.69 & 68.37  & 86.98  & 92.44  \\
			\multicolumn{2}{c}{+ DFFC} & \textbf{99.86} & \textbf{92.26} & \textbf{71.75} & \textbf{90.63} & \textbf{93.50} \\
			\midrule
			\multicolumn{2}{c}{SPSL\cite{SpatialPhaseShallowLearning2021liu}} & 99.36  & 76.88  & 61.51  & 80.93  & 91.56  \\
			\multicolumn{2}{c}{+ DIH} & 99.31  & 81.06 & 58.81  & 82.05  & \textbf{92.38} \\
			\multicolumn{2}{c}{+ DFFC} & \textbf{99.68} & \textbf{82.24} & \textbf{62.36} & \textbf{82.10} & 92.23  \\
			\midrule
			\multicolumn{2}{c}{SRM\cite{GeneralizingFaceForgery2021luo}} & 99.68 & 73.01 & 60.79 & 79.11 & 89.46 \\
			\multicolumn{2}{c}{+ DIH} & 99.64 & 79.01 & 57.81 & 79.29 & 91.88 \\
			\multicolumn{2}{c}{+ DFFC} & \textbf{99.74} & \textbf{81.42} & \textbf{61.22} & \textbf{85.99} & \textbf{93.68} \\
			\bottomrule
	\end{tabular}}%
	\label{tab:cross}%
\end{table}%

\section{Experiments}

\subsection{Experiment Settings}

%\begin{figure}[t]
%	\centering
%	\includegraphics[height=0.5\linewidth,width=0.75\linewidth]{image/epoch_DIH.eps}
%	\caption{Average DIH value of samples at different epoch.}
%	\label{fig:DIH}
%\end{figure}

\noindent \textbf{Datasets and pre-processing.}
In this paper, we mainly conducted experiments on the FaceForensics++ (FF++) dataset\cite{FaceForensicsLearningDetect2019rossler}.
% It contains 1000 Pristine (PT) videos (i.e., the real sample) and 5000 fake videos forged by five manipulation methods, i.e., Deepfakes (DF), Face2Face (F2F), FaceSwap (FS), NeuralTextures (NT) and FaceShifter (FSh). 
% Besides, FF++ provides three quality levels in compression for these videos: raw, high-quality (HQ) and low-quality (LQ). 
The samples were split into disjoint training, validation, and testing sets at the video level follows the official protocol. As for pre-processing, we utilized MTCNN to detect and crop the face regions (enlarged by a factor of 1.3) from each video frame, and resized the them to 256 $\times$ 256.

\noindent \textbf{Implementation detail.}
%We conducted all experiments with PyTorch on a workstation equipped with four NVIDIA Tesla A100 GPUs (40GB memory). 
We employed a SGD optimizer with a cosine learning rate scheduler with $\eta_{max}=0.1$. As for DFFC, we set 20 epoch for totally training, the pacing sequence $\mathbf{T}=[2,5,8,12,15]$ and hyper-parameters $\gamma=0.9,\alpha_{f}=0.5,\alpha_{k}=0.9, E=1000$.

%\subsection{Experimental Analysis}
%In this section, we evaluate our method with several method, including . Considering the fair compassion, with the public . we train our models on FF++. All experimental results are reported in terms of ACC (accuracy) and AUC (area under the curve). 
%We sample all frames for each video to calculate the frame-level ACC and AUC.

% Table generated by Excel2LaTeX from sheet 'Sheet1'
\begin{table}[htbp]
  \centering
  \caption{Video-level AUC(\%) on cross-manipulation evaluations. The best results are highlighted.}
    
    \begin{tabular}{cccccr}
    \toprule
    Method & F2F   & FS    & NT    & FSh   & \multicolumn{1}{c}{Avg} \\
    \midrule
    Xception & 71.13  & 54.87  & 76.95  & 74.72  & 69.42  \\
    +DIH   & 76.70  & 56.50  & 76.90  & 68.14  & 69.56  \\
    +DFFC  & \textbf{84.34 } & \textbf{62.06 } & \textbf{78.70 } & \textbf{75.14 } & \textbf{75.06 } \\
    \bottomrule
    \end{tabular}%
  \label{tab:on cross-manipulation}%
\end{table}%

\subsection{Evaluation of Detection Performances}
%\noindent \textbf{Benchmark deepfake detectors.}
In this part, we deployed the proposed DFFC on several end-to-end deepfake detectors for evaluation. We considered three typical kinds of detection methods:  \textit{1) Spatial detector}, i.e., Xception\cite{XceptionDeepLearning2017chollet}, EfficientNet (ENb4)\cite{EfficientNetRethinkingModel2019tan}, SwinTransformerV2 (Swin) \cite{SwinTransformerV22022liu} and MAT\cite{MultiAttentionalDeepfakeDetection2021zhao}; \textit{2) Frequency detector}, i.e.,  SPSL\cite{SpatialPhaseShallowLearning2021liu}; 3) \textit{Detector contained spatial and frequency branches}, i.e., SRM\cite{GeneralizingFaceForgery2021luo}. We reproduced the aforementioned methods by their official codes and initialized them with imagenet weights. 
%we train our models on FF++. 
%All experimental results are reported in terms of ACC (accuracy) and AUC (area under the curve). 

\noindent \textbf{Within datasets evaluations.}
 We evaluated the performance on detecting five manipulation methods on FF++ (LQ).
As shown in Table \ref{tab:LQ}, the proposed DFFC can improve performance for all benchmarks.

\noindent \textbf{Cross datasets evaluations.}
We evaluated the generalization capability of the proposed DFFC by training on FF++/DF (HQ) and testing on several benchmark datasets, i.e., Celeb-DF (CDF)\cite{CelebDFLargeScaleChallenging2020li} WildDeepfake(Wild)\cite{WildDeepfakeChallengingRealWorld2020zi}, Deepfake Detection (DFD)\footnote{http://ai.googleblog.com/2019/09/contributing-data-to-deepfake-detection.html} and Deepfake Detection Challenge preview (DFDC-p)\cite{DeepfakeDetectionChallenge2019dolhansky}. 
As shown in Table \ref{tab:cross}, our DFFC can improve detection performance for all benchmarks.

\noindent \textbf{Cross manipulation evaluations.}
We conducted the cross-manipulation experiment on FF++, where our model was trained on the DF(HQ) subset and tested on the remaining four manipulations. Take Xception as an example, the result is shown in Table \ref{tab:on cross-manipulation}, our DFFC can improve
detection performance for all benchmarks.

\noindent \textbf{Ablation studies of facial quality hardness.} 
In this part, we investigated the impact of facial quality hardness. We only used DIH\cite{CurriculumLearningDynamic2020zhou}, which removes facial quality hardness $q(i)$ compared to DFFC, to train detectors. As shown in Table \ref{tab:LQ} and Table \ref{tab:cross}, we can observe that only introducing DIH still improve both within and cross datasets performance for most benchmarks, while introducing the facial quality hardness improve further. It demonstrates that both introducing dynamic hardness and facial quality priors are beneficial to training deepfake detectors.
%This is because the introduction of FQS can better screen out difficult samples. Through a more detailed definition of hardness, the model can have better curriculum learning, which makes the model better generalization.

\begin{figure}[t]
	\centering
	\includegraphics[height=0.48\linewidth,width=0.75\linewidth]{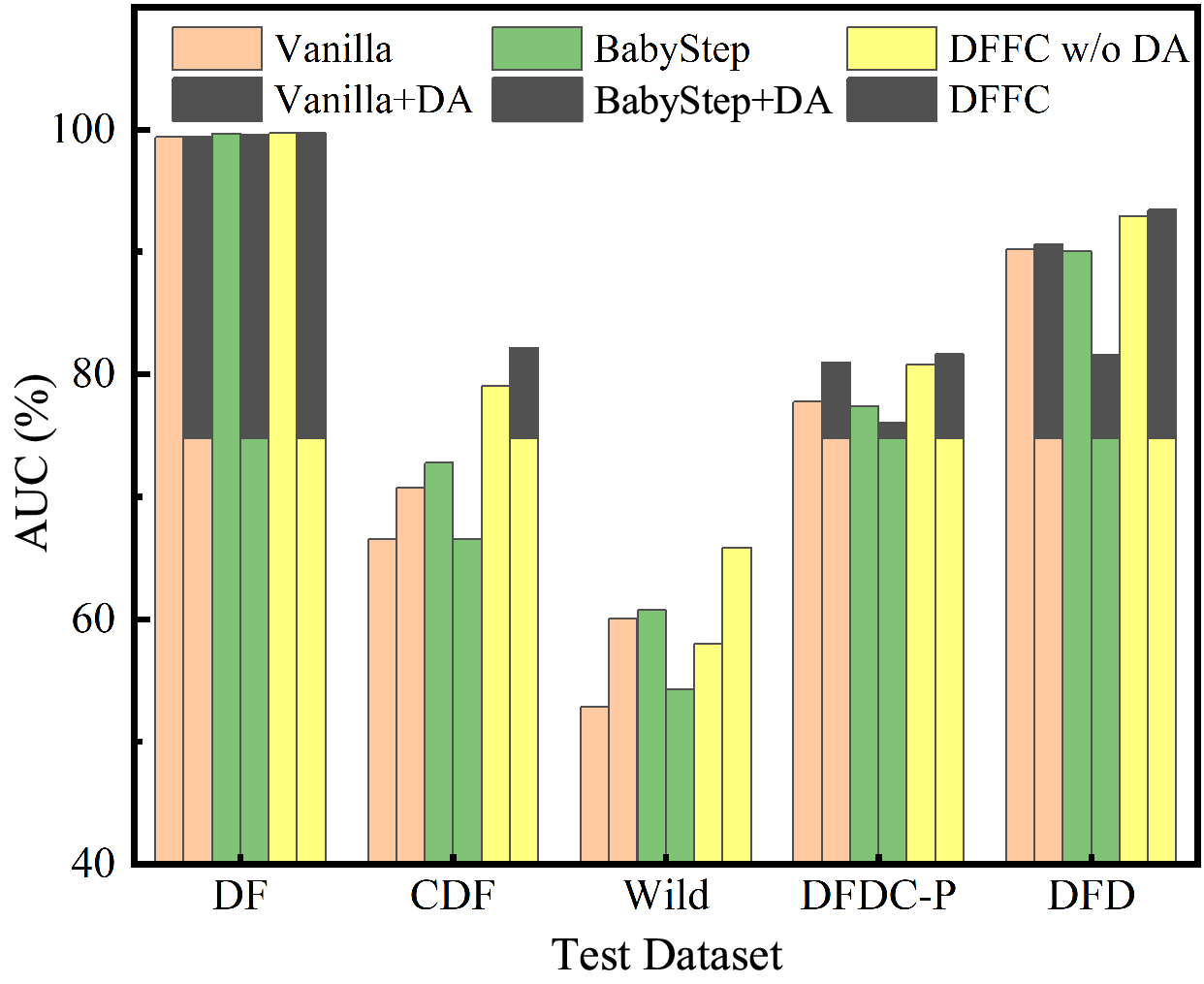}
	\caption{Cross-datasets performance of different training strategies. Trained on FF++/DF(HQ) with Xception.}
	\label{fig:ablation}
\end{figure}

\noindent \textbf{Comparison with other training strategies.} In this part, we compared DFFC with other training strategies, including the vanilla training and BabyStep\cite{CurriculumLearning2009bengio,PowerCurriculumLearning2019hacohen,cirik2016visualizing} which is the simplest CL strategy that utilizes a static pre-defined hardness. We conducted BabyStep by introducing IFQA\cite{IFQAInterpretableFace2023jo} score as the pre-defined hardness and utilizing the pacing setting in \cite{PowerCurriculumLearning2019hacohen}. Furthermore, we also investigated impacts of data augmentations.
As shown in Figure \ref{fig:ablation}, we observe that both CL paradigm and data augmentations can improve performance compared with vanilla training in most cases. 
However, introducing data augmentations in BabyStep suffers severe performance degradation, as it makes the augmented data does not match its pre-defined static hardness.
It demonstrates that the dynamic CL strategy with data augmentations (i.e., our DFFC) is beneficial for general deepfake detection.

\begin{figure}[t]
	\centering
	\includegraphics[height=0.485\linewidth, width=0.75\linewidth]{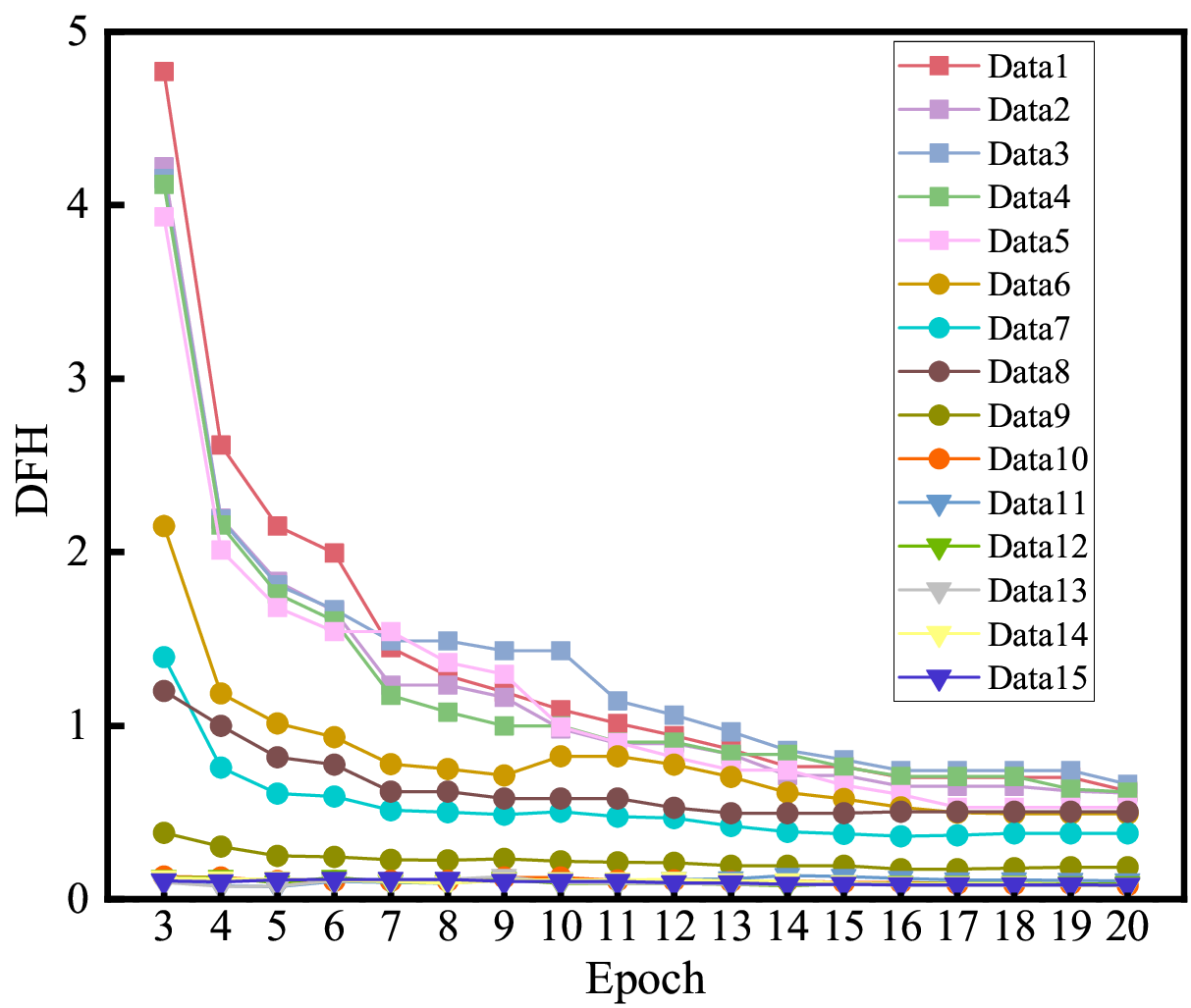}
	\caption{Variations of DFH during the training. We illustrate the change of 5 highest (Data 1-5), 5 lowest (Data 11-15), and random 5 median (Data 6-10) DFH samples from 3rd epoch.}
	\label{fig:sample_DIH}
\end{figure}

\begin{figure}[t]
	\centering
	\subfigure[Real faces with top DFH]{
		\includegraphics[width=0.47\linewidth]{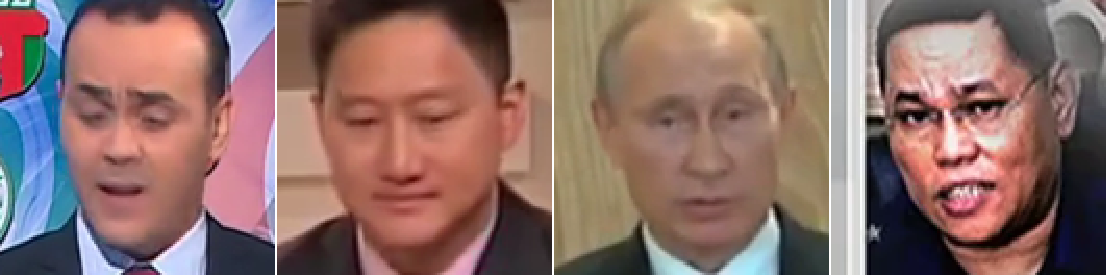}
		\label{fig:sub1}
	}
	\subfigure[Real faces with bottom DFH]{
		\includegraphics[width=0.47\linewidth]{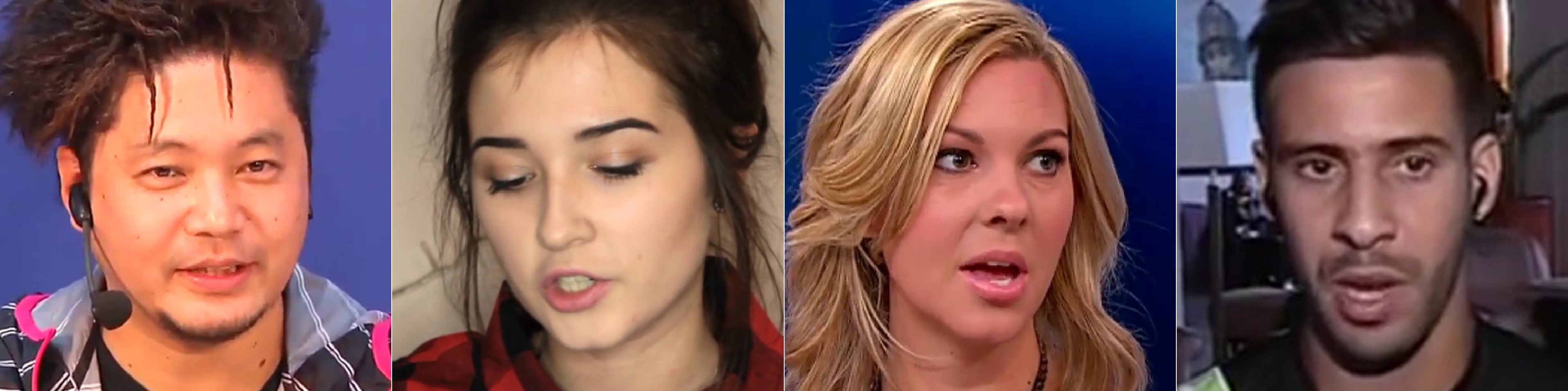}
		\label{fig:sub2}
	}
	
	\subfigure[Fake faces with top DFH]{
		\includegraphics[width=0.47\linewidth]{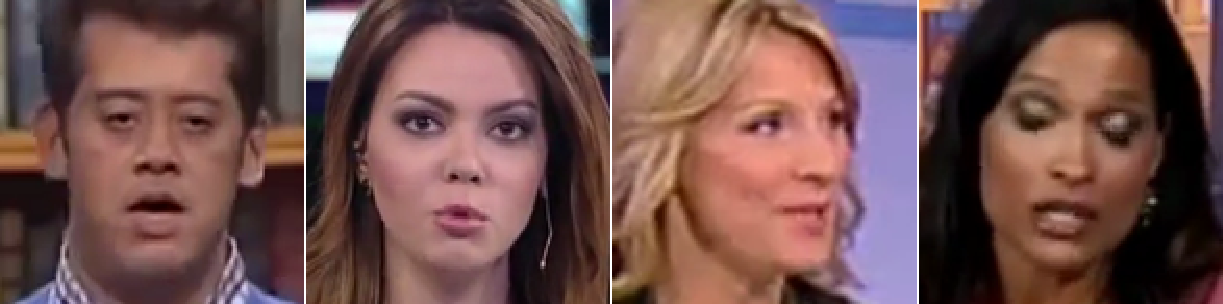}
		\label{fig:sub3}
	}
	\subfigure[Fake faces with bottom DFH]{
		\includegraphics[width=0.47\linewidth]{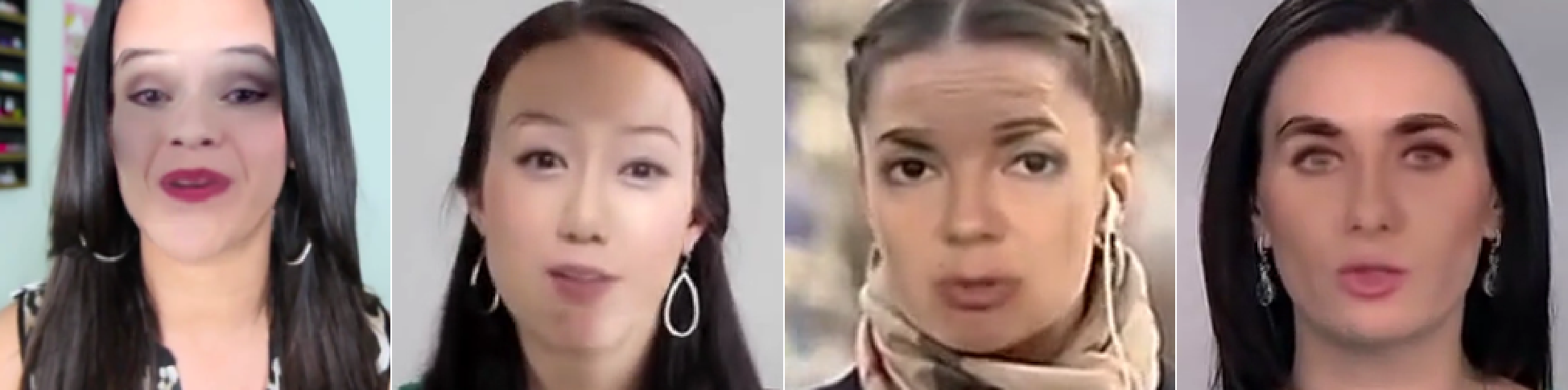}
		\label{fig:sub4}
	}
	\caption{Visualizations of top and bottom DFH for real and fake faces on FF++/DF(HQ). 
		%		A larger DFH indicates that the sample is harder for the training.
	}
	\label{fig:vis}
\end{figure}

\begin{table}[t]
	\centering
	\caption{Average metrics of different DFH samples.}
	\setstretch{0.85}
	\resizebox{\linewidth}{!}{
		\begin{tabular}{ccccccc}
			\toprule
			Metric & DFH-idx    & DF    & F2F   & FS    & NT    & FSh \\
			\midrule
			\multirow{2}[2]{*}{TAR(\%)} & Top   & 2.72  & 1.13  & 1.17  & 3.25  & 2.48 \\
			& Bottom  & 6.84 & 8.26 & 11.31 & 5.40 & 10.84 \\
			\midrule
%			\multirow{2}[2]{*}{FIQ} & Top   & 0.233 & 0.342 & 0.370 & 0.194 & 0.297 \\
%			& Bottom  & \textbf{0.301} & \textbf{0.530} & \textbf{0.562} & \textbf{0.532} & \textbf{0.407} \\
			\multirow{2}[2]{*}{SSIM} & Top   & 0.9874  & 0.9920  & 0.9915  & 0.9803  & 0.9886  \\
			& Bottom  &  0.9012 & 0.9489 & 0.9376 & 0.9674 & 0.9334 \\
			\bottomrule
	\end{tabular}}%
	\label{tab:quality}%
\end{table}%

\subsection{Analysis Properties of DFFC}
In this part, we analyze some properties of DFFC. We conduct the subsequent experiments by utilizing Xception trained on FF++/DF(HQ). 

\noindent \textbf{How DFH changes during the training?}
In this part, we explored the properties of the training process of DFFC.  We respectively selected samples with top (hard samples), bottom (easy samples) and median DFH and illustrated the variations of their DFH during the training.
As shown in Figure \ref{fig:sample_DIH}, we can observe that DFH decreases along the training for all samples. We can also find that DFH of easy samples remains small throughout training. 
This is because deepfake detectors can learn to identify easy samples in the early stages of training so that DFFC does not tend to update their DFH. However, the detectors needs more time to mine the forged clues of the difficult samples.
It indicates that as learning continues, easy samples become less informative so that we can select and train on fewer hard samples.

\noindent \textbf{What do hard samples look like?}
We explored the properties of samples with different hardness mined by DFFC. We respectively illustrated the samples with top and bottom DFH in Figure \ref{fig:vis}. We can observe that the fake faces with top DFH are relatively high-quality compared to those with  bottom DFH, which cannot be easily distinguished. These fake samples with bottom DFH always contain more clear forgery clues, such as color inconsistency. For real faces, top DFH samples usually have heavy post-processing which is easy to confuse with fake faces. It demonstrates that the DFH score mined by our DFFC is in line with human visual perception. 
We also computed the pixel-level tampering ratio (TAR) and SSIM\cite{ImageQualityAssessment2004wang} metrics for fake samples with their corresponding real faces. As shown in the Table \ref{tab:quality}, fake faces with top DFH involve low TAR and high SSIM, which indicates these samples have less forgery artifacts and higher similarities to the corresponding real faces. It makes sense that deepfake detectors would have difficulty identifying these samples.
\section{Conclusion}
\linespread{1}
In this paper, we propose DFFC as a general training strategy for deepfake detection.
First, we present DFH, an innovative metric that integrates the facial quality score and instantaneous instance loss to dynamically evaluate the sample hardness during the training process.  Furthermore, we introduce  a pacing function  that controls the  training subsets from easy to hard basd on  DFH  throughout the iterations of training. It makes deepfake detectors gradually focus on hard samples to mine the general forgery clues during the training.
% Comprehensive experiments show that DFFC can improve both within- and cross-dataset performance of various kinds of end-to-end deepfake detectors through a plug-and-play way. 
%The proposed method can be further extended and applied to other machine learning tasks to improve model performance.
%
\small
\vfill\pagebreak

\footnotesize
\bibliographystyle{IEEEbib}
 \bibliography{icassp_refs}
\end{document}